%% file: main.tex
\pdfoutput=1

\documentclass[11pt]{article}

\usepackage{acl}

\usepackage{times}
\usepackage{latexsym}
\usepackage{CJKutf8}
\usepackage{makecell}

\usepackage[T1]{fontenc}

\usepackage[utf8]{inputenc}

\usepackage{microtype}

\usepackage{amsmath}
\usepackage{amssymb}
\usepackage{xcolor}
\usepackage{arydshln}
\usepackage{tikz}
\usepackage[figuresright]{rotating}
\usetikzlibrary{arrows, decorations.text, shapes.geometric, positioning, decorations.pathreplacing, calligraphy}
\usepackage{pgfplots}
\usepackage{pgfmath}
\usepackage{pgffor}
\pgfplotsset{compat=1.17}
\usepackage{subcaption}
\usepackage{mathtools}
\usepackage{multirow}
\usepackage{booktabs}
\usepackage{pifont}
\usepackage{ifthen}

\title{Alibaba-Translate China's Submission for\\WMT 2022 Metrics Shared Task}

\author{
    Yu Wan$^{1,2}$\thanks{~~Equal contribution. Work was done when Yu Wan and Keqin Bao were interning at DAMO Academy, Alibaba Group.}~~~Keqin Bao$^{1,3*}$~~~Dayiheng Liu$^1$~~~Baosong Yang$^1$~~~\textbf{Derek F. Wong}$^2$\\\textbf{Lidia S. Chao}$^2$~~~\textbf{Wenqiang Lei}$^4$~~~\textbf{Jun Xie}$^1$\\
    $^1$DAMO Academy, Alibaba Group~~~~~~$^2$NLP$^2$CT Lab, University of Macau\\
    $^3$University of Science and Technology of China~~~~~~$^4$National University of Singapore\\
    \vspace{-0.75ex}
    \small{\tt{nlp2ct.ywan@gmail.com~~~baokq@mail.ustc.edu.cn}}\\
    \vspace{-0.75ex}
    \small{\tt{\{liudayiheng.ldyh,yangbaosong.ybs,qingjing.xj\}@alibaba-inc.com}}\\
    \vspace{-0.75ex}
    \small{\tt{\{derekfw,lidiasc\}@um.edu.mo}~~~\tt{wenqianglei@gmail.com}}
}

\begin{document}
\maketitle



\input{abstract}
\input{introduction}

\input{methods}

\input{experiments}
\input{table_main}

\input{analysis}

\input{conclusion}

\section*{Acknowledgements}

The participants would like to send great thanks to the committee and the organizers of the WMT Metrics Shared Task competition.
Besides, the authors would like to thank the reviewers and meta-review for their insightful suggestions.

This work was supported in part by the Science and Technology Development Fund, Macau SAR (Grant No. 0101/2019/A2), the Multi-year Research Grant from the University of Macau (Grant No. MYRG2020-00054-FST), National Key Research and Development Program of China (No. 2018YFB1403202), and Alibaba Group through Alibaba Innovative Research (AIR) Program.

\bibliographystyle{acl_natbib}
\bibliography{anthology}

\end{document}

%% file: abstract.tex
\begin{abstract}
In this report, we present our submission to the WMT 2022 Metrics Shared Task.
We build our system based on the core idea of~\textsc{\textbf{UniTE}} (\textbf{Uni}fied \textbf{T}ranslation \textbf{E}valuation), which unifies source-only, reference-only, and source-reference-combined evaluation scenarios into one single model.
Specifically, during the model pre-training phase, we first apply the pseudo-labeled data examples to continuously pre-train~\textsc{UniTE}.
Notably, to reduce the gap between pre-training and fine-tuning, we use data cropping and a ranking-based score normalization strategy.
During the fine-tuning phase, we use both Direct Assessment (DA) and Multidimensional Quality Metrics (MQM) data from past years' WMT competitions.
Specially, we collect the results from models with different pre-trained language model backbones, and use different ensembling strategies for involved translation directions.

\end{abstract}

%% file: introduction.tex
\section{Introduction}
\label{sec.intro}
Translation metric aims at delivering accurate and convincing predictions to identify the translation quality of outputs with access to one or many gold-standard reference translations~\cite{ma-etal-2018-results,ma-etal-2019-results,mathur-etal-2020-results,freitag-etal-2021-results}.
As the development of neural machine translation research~\cite{vaswani2017attention,wei-etal-2022-learning}, the metric methods should be capable of evaluating the high-quality translations at the level of semantics rather than surfance-level features~\cite{sellam-etal-2020-bleurt,ranasinghe-etal-2020-transquest,rei-etal-2020-comet,wan-etal-2022-unite}.
In this paper, we describe Alibaba Translate China's submissions to the WMT 2022 Metrics Shared Task to deliver a more adequate evaluation solution at the level of semantics.

Pre-trained language models (PLMs) like BERT~\cite{devlin-etal-2019-bert} and XLM-R~\cite{conneau-etal-2020-unsupervised} have shown promising results in identifying the quality of translation outputs.
Compared to conventional statistical- (\textit{e.g.}, BLEU,~\citealp{papineni-etal-2002-bleu}  and representation-based methods (\textit{e.g.},~\textsc{BERTScore},~\citealp{zhang2019bertscore}), the model-based approaches (\textit{e.g.},~\citealp[BLEURT,][]{sellam-etal-2020-bleurt};~\citealp[COMET,][]{rei-etal-2020-comet};~\citealp[\textsc{UniTE},][]{wan-etal-2022-unite}) show their strong ability on delivering more accurate quality predictions, especially those approaches which apply source sentences as additional input for the metric model~\cite{rei-etal-2020-comet,takahashi-etal-2020-automatic,wan-etal-2021-robleurt,wan-etal-2022-unite}.
Specifically, those metric models are designed as a combination of PLM and feedforward network, where the former is in charge of deriving representations on input sequence, and the latter predicts the translation quality based on the representation.
The metric model, which is trained on synthetic or human annotations following a regressive objective, learns to mimic human predictions to identify the translation quality of the hypothesis sentence.

Although those model-based metrics have shown promising results in modern applications and translation quality estimation, they still show their own shortcomings as follows.
First, they often handle one specific evaluation scenario,~\textit{e.g.}, COMET serves source-reference-only evaluation, where the source and reference sentence should be concurrently fed to the model for prediction.
For the other evaluation scenarios, they hardly give accurate predictions, showing the straits of metric models due to the disagreement between training and inference.
Besides, recent studies have investigated the feasibility of unifying those evaluation scenarios into one single model, which can further improve the evaluation correlation with human ratings in any scenario among source-only, reference-only, and source-reference-combined evaluation~\cite{wan-etal-2021-robleurt,wan-etal-2022-unite}.
This indicates that, training with multiple input formats than a specific one can deliver more appropriate predictions for translation quality identification.
More importantly, unifying all translation evaluation functionalities into one single model can serve as a more convenient toolkit in real-world applications.

Following the idea of~\newcite{wan-etal-2022-unite} and the experience in previous competition~\cite{wan-etal-2021-robleurt}, we directly use the pipeline of \textsc{UniTE}~\cite{wan-etal-2022-unite} to build models for this year's metric task.
Each of our models can integrate the functionalities of source-only, reference-only, and source-reference-combined translation evaluation into itself.
When collecting the system outputs for the WMT 2022 Metrics Shared Task, we employ our~\textsc{UniTE} models to predict the translation quality scores following the source-reference-combined setting.
Compared to the previous version of~\textsc{UniTE}~\cite{wan-etal-2022-unite}, we reform the synthetic training set for the continuous pre-training phase, raising the ratio of training examples consisting of high-quality hypothesis sentences.
Also, during fine-tuning our metric model, we apply available Direct Assessment~\citep[DA,][]{bojar-etal-2017-results,ma-etal-2018-results,ma-etal-2019-results,mathur-etal-2020-results} and Multidimensional Quality Metrics datasets~\citep[MQM,][]{freitag-etal-2021-experts,freitag-etal-2021-results} from previous WMT competitions to further improve the performance of our model.
Specifically, for each translation direction among English to German (En-De), English to Russian (En-Ru), and Chinese to English (Zh-En) directions, we applied different ensembling strategies to achieve a better correlation with human ratings on MQM 2021 dataset.
Results on WMT 2021 MQM dataset further demonstrate the effectiveness of our method.

%% file: methods.tex
\section{Method}
As outlined in \S\ref{sec.intro}, we apply the~\textsc{UniTE} framework~\cite{wan-etal-2022-unite} to obtain metric models.
We use three types of input formats (\textit{i.e.}, source-only, reference-only, and source-reference-combined) during training. 
While during inference, we only use the source-reference-combined paradigm to collect evaluation scores.
In this section, we introduce the applied model architecture (\S\ref{subsec.model_architecture}), synthetic data construction method (\S\ref{subsec.synthetic_data_construction}), and model training strategy (\S\ref{subsec.training_pipeline}) for this year's metric competition.

\subsection{Model architecture}
\label{subsec.model_architecture}

\paragraph{Input Format}
Following~\newcite{wan-etal-2022-unite}, we construct the input sequence for source-only, reference-only, and source-reference-combined input formats as follows:
\begin{align}
    \mathbf{x}_{\text{\textsc{Src}}} &= \texttt{[BOS]} \mathbf{h} \texttt{[DEL]} \mathbf{s} \texttt{[EOS]}, \\
    \mathbf{x}_{\text{\textsc{Ref}}} &= \texttt{[BOS]} \mathbf{h} \texttt{[DEL]} \mathbf{r} \texttt{[EOS]}, \\
    \mathbf{x}_{\text{\textsc{Src+Ref}}} &= \texttt{[BOS]} \mathbf{h} \texttt{[DEL]} \mathbf{s} \texttt{[DEL]} \mathbf{r} \texttt{[EOS]},
\end{align}
where \texttt{[BOS]}, \texttt{[DEL]} and \texttt{[EOS]} represent the beginning, the delimeter, and the ending of sequence,\footnote{Those symbols may vary if we use different PLMs, \textit{e.g.}, ``[BOS]'', ``[SEP]'', and ``[SEP]'' for English BERT~\cite{devlin-etal-2019-bert}, and ``<s>'', ``</s> </s>'', and ``</s>'' for XLM-R~\cite{conneau-etal-2020-unsupervised}.} and $\mathbf{h}$, $\mathbf{s}$, and $\mathbf{r}$ are hypothesis, source, and reference sentence, respectively.
During the pre-training phase, we applied all input formats to enhance the performance of \textsc{UniTE} models.

\paragraph{Model Backbone Selection}
Aside from the reference sentence which is written in the same language as the hypothesis sentence, the source is in another different language.
We believe that, cross-lingual semantic alignments can ease the model training on source-only and source-reference-combined scenarios.
Referring to the setting of existing methods~\cite{ranasinghe-etal-2020-transquest,rei-etal-2020-comet,sellam-etal-2020-bleurt,wan-etal-2022-unite}, they apply XLM-R~\cite{conneau-etal-2020-unsupervised} as the backbone of evaluation models for better multilingual support.
In this competition, we additionally use~\textsc{infoXLM}~\cite{chi-etal-2021-infoxlm}, which enhances the XLM-R model with cross-lingual alignments, as the backbone of our \textsc{UniTE} models.

\paragraph{Model Training}
Following~\newcite{wan-etal-2022-unite}, we first equally split all examples into three parts, each of which only serves one input format training.
As to each training example, after concatenating the required input sentences into one sequence and feeding it to PLM, we collect the corresponding representations --  $\mathbf{H}_\textsc{Ref}, \mathbf{H}_\textsc{Src}, \mathbf{H}_\textsc{Src+Ref}$ for each input format, respectively.
After that, we use the output embedding assigned with CLS token $\mathbf{h}$ as the sequence representation.
Finally, a feedforward network takes $\mathbf{h}$ as input and gives a scalar $p$ as a prediction.
Taking $\mathbf{x}_{\textsc{Src}}$ as an example: 
\begin{align}
    \mathbf{H}_\textsc{Src} &= \texttt{PLM}(\mathbf{x}_\textsc{Src}) \in \mathbb{R}^{ (l_h + l_s) \times d}, \\
    \mathbf{h}_\textsc{Src}& = \texttt{CLS}(\mathbf{H}_\textsc{Src}) \in \mathbb{R}^{d}, \\
    p_\textsc{Src} & = \texttt{FeedForward}(\mathbf{h}_\textsc{Src}) \in \mathbb{R}^{1},
\end{align}
where $l_h$ and $l_s$ are the lengths of $\mathbf{h}$ and $\mathbf{s}$, respectively.

For learning objectives, we apply the mean squared error (MSE) as the loss function:
\begin{align}
   \mathcal{L}_\textsc{SRC} = (p_\textsc{SRC} - q) ^ 2,
\end{align}
where $q$ is the given ground-truth score.
Note that, the batch size is the same across all input formats to avoid the training imbalance.
During each update, the final learning objective is the sum of losses for all formats:
\begin{align}
    \mathcal{L} = \mathcal{L}_\textsc{Ref} + \mathcal{L}_\textsc{Src} + \mathcal{L}_\textsc{Src+Ref}.
\end{align}

\subsection{Synthetic Data Construction}
\label{subsec.synthetic_data_construction}
To better enhance the translation evaluation ability, we first construct a synthetic dataset for continuous pre-training.
The overall stage for obtaining the dataset consists of the following steps:
1) collecting synthetic data from parallel data provided by the WMT Translation task;
2) downgrading the translation quality and keeping the consistency of synthetic and MQM datasets;
3) relabeling them with a ranking-based scoring strategy.

\paragraph{Collecting Synthetic Data}
Specifically, we first conduct parallel data from this year's WMT Translation competition as the source-reference sentence pairs.
Then, we obtain hypothesis sentences via translating the source using online translation engines,~\textit{e.g.},~\texttt{Google Translate}\footnote{\href{https://translate.google.com}{\url{https://translate.google.com}}} and~\texttt{Alibaba Translate}\footnote{\href{https://translate.alibaba.com}{\url{https://translate.alibaba.com}}}.

\paragraph{Quality Downgrading}
We follow existing works~\cite{sellam-etal-2020-bleurt,wan-etal-2022-unite} to apply the word/span dropping strategy to downgrade the quality of hypothesis sentences, thus increasing the ratio of training examples consisting of bad translation outputs.
Specially, we notice that the translation quality of hypothesis sentences in the MQM dataset is rather higher than that in the DA dataset.
In practice, to reduce the translation quality distribution gap between the synthetic and MQM datasets, we randomly select 15\% examples of the entire dataset, which is lower than the applied ratio (\textit{i.e.}, 30\%) in BLEURT~\cite{sellam-etal-2020-bleurt} and \textsc{UniTE}~\cite{wan-etal-2022-unite}.

\paragraph{Data Labeling}
After downgrading the translation quality of synthetic hypothesis sentences, we then collect predicted scores for each triple as the learning supervision.
To increase the confidence of pseudo-labeled scores, we use multiple~\textsc{UniTE} checkpoints trained with different random seeds to label the synthetic data.
Besides, to reduce the gap of predicted scores among different translation directions, we applied the ranking-based scoring strategy as in~\newcite{wan-etal-2022-unite}.

\subsection{Training Pipeline}
\label{subsec.training_pipeline}

\paragraph{Pre-train with Synthetic Data}
First, we use the synthetic dataset to continuously pre-train our~\textsc{UniTE} models to enhance the evaluation ability on three input formats.

\paragraph{Fine-tune with DA Dataset}
After training \textsc{UniTE} models on the synthetic dataset, we apply the DA dataset for the first stage of model fine-tuning.
Considering the support of multilingual translation evaluation, we collect all DA datasets from the previous years, and we leave the year 2021 out of training due to the reported bug from the official committee.
We think that, although the DA and MQM datasets have different scoring rules, training~\textsc{UniTE} models on DA as an additional phase can enhance both the model robustness and the support of multilingualism.
Besides, the number of examples in the DA dataset is extremely larger than that in MQM.
The training examples from the DA dataset can provide more learning signals for~\textsc{UniTE} model training.

\paragraph{Fine-tune with MQM Dataset}
After fine-tuning \textsc{UniTE} models on the DA dataset, we then apply the MQM dataset for the second stage of model fine-tuning.
For this year's competition, we first use MQM 2020 dataset during this stage, and testify the performance of our models on MQM 2021 to tune the hyper-parameters.
Then, after identifying the hyper-parameters, we use all MQM datasets to fine-tune, choose two models whose backbones are XLM-R and~\textsc{infoXLM}, and collect the ensembled scores as submissions.

\subsection{Model Ensembling}
For each training pipeline, we use the three random seeds to train~\textsc{UniTE} models.
However, when identifying the performance of all models on the MQM 2021 dataset, we find it hard to select the same strategy across all domains and translation directions.
In practice, we select the models trained with different random seeds for each translation direction.

%% file: experiments.tex
\section{Experiments}
\subsection{Experiment Settings}

\paragraph{Implementations}
All of our models are implemented with the released \textsc{UniTE} repository.\footnote{\href{https://github.com/wanyu2018umac/UniTE}{\url{https://github.com/wanyu2018umac/UniTE}}}
We choose the large version of XLM-R~\cite{conneau-etal-2020-unsupervised} and~\textsc{infoXLM}~\cite{chi-etal-2021-infoxlm} as the PLM backbones of all~\textsc{UniTE} models, and directly use the released checkpoints from Huggingface Transformers~\cite{wolf-etal-2020-transformers}.\footnote{\href{https://huggingface.co/xlm-roberta-large}{\url{https://huggingface.co/xlm-roberta-large}}, \href{https://huggingface.co/microsoft/infoxlm-large}{\url{https://huggingface.co/microsoft/infoxlm-large}}}

\paragraph{Continuous pre-training}
Following~\newcite{wan-etal-2022-unite}, we collect the translation hypotheses from 10 directions,~\textit{i.e.}, English-Czech/German/Japanese/Russian/Chinese, as those translation directions are engaged with massive parallel datasets and the performance of corresponding online translation engines is relatively high.
For each translation direction, we collect 0.5M hypotheses, and label the translation quality scores as describled in \S\ref{subsec.synthetic_data_construction}.

\paragraph{Hyper-parameters}
Following the setting in~\newcite{wan-etal-2022-unite}, the feedforward network of our \textsc{UniTE} model contains three linear transition layers, whose output dimensionalities are 3,072, 1,024, and 1, respectively.
Between any two adjacent layers, the hyperbolic tangent is arranged as the activations.
During the continuous pre-training phase, we set the batch size for each input format as 1,024, and tune the hyper-parameters for our models.
For the models whose backbone is XLM-R, the learning rates for PLM and feedforward network are $1.0 \cdot 10^{-4}$ for PLM, and $3.0 \cdot 10^{-4}$, respectively.
For the models whose backbone is~\textsc{infoXLM}, the learning rates are $5.0 \cdot 10^{-5}$ for PLM, and $1.5 \cdot 10^{-4}$, respectively.
For all the fine-tuning steps, we use the batch size as 32 across all settings, and the learning rates for PLM and feedforward network are $5.0 \cdot 10^{-6}$ for PLM, and $1.5 \cdot 10^{-5}$, respectively.

\paragraph{Performance Evaluation}
Following the previous setting ~\cite{ma-etal-2018-results,ma-etal-2019-results,mathur-etal-2020-results,freitag-etal-2021-results}, we use the variant Kendall's Tau to evaluate the performance of our models on the MQM 2021 dataset.
For comparison, we directly use the officially released COMET checkpoints~\cite{rei-etal-2020-comet}\footnote{\href{https://github.com/Unbabel/COMET/}{\url{https://github.com/Unbabel/COMET/}}}, and select the checkpoints which are trained with DA or MQM datasets.

\paragraph{Results Conduction}
When collecting the results for submitting predictions, we ensembled the models by directly averaging the predictions on the same example.
We do not apply the idea of uncertainty-aware sampling~\cite{zhou-etal-2020-uncertainty,wan-etal-2020-self,glushkova-etal-2021-uncertainty-aware} during inference, because it takes far more additional time to collect the results.

%% file: table_main.tex
\begin{table*}[t]
    \centering
    \scalebox{1.0}
    {
        \begin{tabular}{lccccccc}
            \toprule
            \multirow{2}{*}{\textbf{Model}} & \multicolumn{3}{c}{\textbf{News}} & \multicolumn{3}{c}{\textbf{TED}} & \multirow{2}{*}{\textbf{All}} \\
            \cmidrule(l{2pt}r{2pt}){2-4}
            \cmidrule(l{2pt}r{2pt}){5-7}
             & En-De & En-Ru & Zh-En & En-De & En-Ru & Zh-En & \\
            \midrule
            COMET-QE-DA-2021 & 23.7 & 34.6 & ~~8.3 & 12.3 & 22.5 & ~~8.5 & 14.4 \\
            COMET-DA-2021 & 28.1 & \textbf{43.1} & 15.2 & 20.2 & 28.5 & 15.9 & 21.6 \\
            COMET-QE-MQM-2021 & 26.7 & 33.3 & ~~6.7 & 10.6 & 22.3 & ~~5.5 & 12.8\\
            COMET-MQM-2021 & 27.5 & 42.5 & 11.4 & 18.5 & 28.8 & 13.3 & 20.2 \\
            
            \cdashline{1-8}[1pt/2.5pt]\noalign{\vskip 0.5ex}

            \textsc{UniTE}$_{\text{XLM-R}}$ & 27.7 & 39.0 & \textbf{16.3} & 19.7 & \textbf{31.2} & \textbf{17.3} & \textbf{25.3} \\
            \textsc{UniTE}$_{\text{\textsc{infoXLM}}}$ & \textbf{40.0} & 36.2 & 13.0 & \textbf{25.3} & 28.7 & ~~9.2 & 24.9 \\

            \bottomrule
                 
        \end{tabular}
    }
    \caption{Kendall's Tau correlation (\%) on MQM 2021 dataset. The best results for each translation direction are bold. Taking XLM-R as backbone shows better result on En-Ru and Zh-En, and~\textsc{infoXLM} on En-De.}
    \label{table_main}
\end{table*}

%% file: analysis.tex
\section{Results and Analysis}
\paragraph{Baselines}
The experimental results are conducted in Table~\ref{table_main}.
As seen, among all involved baselines, the source-only evaluation models (models marked with ``QE'') perform worse than their corresponding source-reference-combined ones, dropping 7.2 and 7.4 Kendall's Tau correlation on DA and MQM settings.
This verifies that, the reference sentence in model translation quality evaluation offers more information for metric models to help deliver accurate predictions~\cite{rei-etal-2020-comet,takahashi-etal-2020-automatic,wan-etal-2022-unite}.
Besides, the model fine-tuned on the DA dataset performs slightly better than that on MQM.
We think that the DA dataset may show its advantage in the robustness of multilingual support and the scale of the training dataset.

\paragraph{\textsc{UniTE} models}
As to our \textsc{UniTE} models, replacing the XLM-R backbone with~\textsc{infoXLM} PLM for metric models does not deliver consistent improvement on average.
Specifically, for both News and TED domains, the \textsc{UniTE} model with~\textsc{infoXLM} as the backbone shows a better correlation on En-De direction, whereas worse on En-Ru and Zh-En than XLM-R.
In addition, the COMET-DA-2021 performs best in En-Ru direction, where we think the reason lies in the scarcity of En-Ru training examples in MQM.
In practice, during collecting the ensembled outputs, we mainly use the~\textsc{UniTE}$_{\text{\textsc{infoXLM}}}$ models for En-De, and~\textsc{UniTE}$_{\text{XLM-R}}$ for En-Ru and Zh-En.

%% file: conclusion.tex
\section{Conclusion}
In this paper, we describe our submission \textsc{UniTE} for the sentence-level Metrics Shared Task at WMT 2022. 
We apply~\textsc{UniTE}~\cite{wan-etal-2022-unite} as the pipeline of our models.
During training, we utilize three input formats to train our models on our synthetic, DA, and MQM data sequentially.
Besides, we ensemble the two models which consist of two different backbones -- \textsc{XLM-R} and \textsc{infoXLM}.
Experiments demonstrate the reliability of our model for identifying the quality of translation outputs, whereas the two models whose backbones \textsc{XLM-R} and \textsc{infoXLM} show different performance for different translation directions.

For the future work, we think that exploring the feasibility of model-based evaluation metrics for other natural language processing tasks is interesting.
We believe that, building reliable evaluation metrics for translation diversity~\cite{lin-etal-2022-bridging,lin-etal-2021-towards}, domain-specific translation quality~\cite{yao-etal-2020-domain,wan-etal-2022-challenges}, and natural language generation~\cite{liu2022kgr4,yang-etal-2021-pos,yang-etal-2022-gcpg} is also of vital importance for the natural language processing community.

Besides, we also submit the source-only predictions of our models to this year's WMT Quality Estimation Shared Task, achieving 1st place on multilingual and En-Ru, and 2nd place on En-De and Zh-En sub-tracks.
This further demonstrates the effectiveness of our \textsc{UniTE} approach, that unifying all evaluation scenarios into one single model can enhance the model performances on all evaluation tasks.
We believe that, the idea of unifying three kinds of translation evaluation functionalities (\textit{i.e.}, source-only, reference-only, and source-reference-combined) into one single model can deliver strong evaluation models on all scenarios.
This research topic is worth further exploration in the future.

%% file: main.bbl
\begin{thebibliography}{30}
\expandafter\ifx\csname natexlab\endcsname\relax\def\natexlab#1{#1}\fi

\bibitem[{Bojar et~al.(2017)Bojar, Graham, and
  Kamran}]{bojar-etal-2017-results}
Ond{\v{r}}ej Bojar, Yvette Graham, and Amir Kamran. 2017.
\newblock \href {https://doi.org/10.18653/v1/W17-4755} {Results of the {WMT}17
  metrics shared task}.
\newblock In \emph{Proceedings of the Second Conference on Machine
  Translation}, pages 489--513, Copenhagen, Denmark. Association for
  Computational Linguistics.

\bibitem[{Chi et~al.(2021)Chi, Dong, Wei, Yang, Singhal, Wang, Song, Mao,
  Huang, and Zhou}]{chi-etal-2021-infoxlm}
Zewen Chi, Li~Dong, Furu Wei, Nan Yang, Saksham Singhal, Wenhui Wang, Xia Song,
  Xian-Ling Mao, Heyan Huang, and Ming Zhou. 2021.
\newblock \href {https://doi.org/10.18653/v1/2021.naacl-main.280} {{I}nfo{XLM}:
  An information-theoretic framework for cross-lingual language model
  pre-training}.
\newblock In \emph{Proceedings of the 2021 Conference of the North American
  Chapter of the Association for Computational Linguistics: Human Language
  Technologies}, pages 3576--3588, Online. Association for Computational
  Linguistics.

\bibitem[{Conneau et~al.(2020)Conneau, Khandelwal, Goyal, Chaudhary, Wenzek,
  Guzm{\'a}n, Grave, Ott, Zettlemoyer, and
  Stoyanov}]{conneau-etal-2020-unsupervised}
Alexis Conneau, Kartikay Khandelwal, Naman Goyal, Vishrav Chaudhary, Guillaume
  Wenzek, Francisco Guzm{\'a}n, Edouard Grave, Myle Ott, Luke Zettlemoyer, and
  Veselin Stoyanov. 2020.
\newblock \href {https://doi.org/10.18653/v1/2020.acl-main.747} {Unsupervised
  cross-lingual representation learning at scale}.
\newblock In \emph{Proceedings of the 58th Annual Meeting of the Association
  for Computational Linguistics}, pages 8440--8451, Online. Association for
  Computational Linguistics.

\bibitem[{Devlin et~al.(2019)Devlin, Chang, Lee, and
  Toutanova}]{devlin-etal-2019-bert}
Jacob Devlin, Ming-Wei Chang, Kenton Lee, and Kristina Toutanova. 2019.
\newblock \href {https://doi.org/10.18653/v1/N19-1423} {{BERT}: Pre-training of
  deep bidirectional transformers for language understanding}.
\newblock In \emph{Proceedings of the 2019 Conference of the North {A}merican
  Chapter of the Association for Computational Linguistics: Human Language
  Technologies, Volume 1 (Long and Short Papers)}, pages 4171--4186,
  Minneapolis, Minnesota. Association for Computational Linguistics.

\bibitem[{Freitag et~al.(2021{\natexlab{a}})Freitag, Foster, Grangier,
  Ratnakar, Tan, and Macherey}]{freitag-etal-2021-experts}
Markus Freitag, George Foster, David Grangier, Viresh Ratnakar, Qijun Tan, and
  Wolfgang Macherey. 2021{\natexlab{a}}.
\newblock \href {https://doi.org/10.1162/tacl_a_00437} {Experts, errors, and
  context: A large-scale study of human evaluation for machine translation}.
\newblock \emph{Transactions of the Association for Computational Linguistics},
  9:1460--1474.

\bibitem[{Freitag et~al.(2021{\natexlab{b}})Freitag, Rei, Mathur, Lo, Stewart,
  Foster, Lavie, and Bojar}]{freitag-etal-2021-results}
Markus Freitag, Ricardo Rei, Nitika Mathur, Chi-kiu Lo, Craig Stewart, George
  Foster, Alon Lavie, and Ond{\v{r}}ej Bojar. 2021{\natexlab{b}}.
\newblock \href {https://aclanthology.org/2021.wmt-1.73} {Results of the
  {WMT}21 metrics shared task: Evaluating metrics with expert-based human
  evaluations on {TED} and news domain}.
\newblock In \emph{Proceedings of the Sixth Conference on Machine Translation},
  pages 733--774, Online. Association for Computational Linguistics.

\bibitem[{Glushkova et~al.(2021)Glushkova, Zerva, Rei, and
  Martins}]{glushkova-etal-2021-uncertainty-aware}
Taisiya Glushkova, Chrysoula Zerva, Ricardo Rei, and Andr{\'e} F.~T. Martins.
  2021.
\newblock \href {https://doi.org/10.18653/v1/2021.findings-emnlp.330}
  {Uncertainty-aware machine translation evaluation}.
\newblock In \emph{Findings of the Association for Computational Linguistics:
  EMNLP 2021}, pages 3920--3938, Punta Cana, Dominican Republic. Association
  for Computational Linguistics.

\bibitem[{Lin et~al.(2022)Lin, Yang, Yao, Liu, Zhang, Xie, Zhang, and
  Su}]{lin-etal-2022-bridging}
Huan Lin, Baosong Yang, Liang Yao, Dayiheng Liu, Haibo Zhang, Jun Xie, Min
  Zhang, and Jinsong Su. 2022.
\newblock \href {https://doi.org/10.18653/v1/2022.findings-naacl.200} {Bridging
  the gap between training and inference: Multi-candidate optimization for
  diverse neural machine translation}.
\newblock In \emph{Findings of the Association for Computational Linguistics:
  NAACL 2022}, pages 2622--2632, Seattle, United States. Association for
  Computational Linguistics.

\bibitem[{Lin et~al.(2021)Lin, Yao, Yang, Liu, Zhang, Luo, Huang, and
  Su}]{lin-etal-2021-towards}
Huan Lin, Liang Yao, Baosong Yang, Dayiheng Liu, Haibo Zhang, Weihua Luo, Degen
  Huang, and Jinsong Su. 2021.
\newblock \href {https://doi.org/10.18653/v1/2021.acl-long.310} {Towards
  user-driven neural machine translation}.
\newblock In \emph{Proceedings of the 59th Annual Meeting of the Association
  for Computational Linguistics and the 11th International Joint Conference on
  Natural Language Processing (Volume 1: Long Papers)}, pages 4008--4018,
  Online. Association for Computational Linguistics.

\bibitem[{Liu et~al.(2022)Liu, Liu, Yang, Zhang, Ding, Yao, Luo, Zhang, and
  Su}]{liu2022kgr4}
Xin Liu, Dayiheng Liu, Baosong Yang, Haibo Zhang, Junwei Ding, Wenqing Yao,
  Weihua Luo, Haiying Zhang, and Jinsong Su. 2022.
\newblock Kgr4: Retrieval, retrospect, refine and rethink for commonsense
  generation.
\newblock In \emph{Proceedings of the AAAI Conference on Artificial
  Intelligence}, volume~36, pages 11029--11037.

\bibitem[{Ma et~al.(2018)Ma, Bojar, and Graham}]{ma-etal-2018-results}
Qingsong Ma, Ond{\v{r}}ej Bojar, and Yvette Graham. 2018.
\newblock \href {https://doi.org/10.18653/v1/W18-6450} {Results of the {WMT}18
  metrics shared task: Both characters and embeddings achieve good
  performance}.
\newblock In \emph{Proceedings of the Third Conference on Machine Translation:
  Shared Task Papers}, pages 671--688, Belgium, Brussels. Association for
  Computational Linguistics.

\bibitem[{Ma et~al.(2019)Ma, Wei, Bojar, and Graham}]{ma-etal-2019-results}
Qingsong Ma, Johnny Wei, Ond{\v{r}}ej Bojar, and Yvette Graham. 2019.
\newblock \href {https://doi.org/10.18653/v1/W19-5302} {Results of the {WMT}19
  metrics shared task: Segment-level and strong {MT} systems pose big
  challenges}.
\newblock In \emph{Proceedings of the Fourth Conference on Machine Translation
  (Volume 2: Shared Task Papers, Day 1)}, pages 62--90, Florence, Italy.
  Association for Computational Linguistics.

\bibitem[{Mathur et~al.(2020)Mathur, Wei, Freitag, Ma, and
  Bojar}]{mathur-etal-2020-results}
Nitika Mathur, Johnny Wei, Markus Freitag, Qingsong Ma, and Ond{\v{r}}ej Bojar.
  2020.
\newblock \href {https://aclanthology.org/2020.wmt-1.77} {Results of the
  {WMT}20 metrics shared task}.
\newblock In \emph{Proceedings of the Fifth Conference on Machine Translation},
  pages 688--725, Online. Association for Computational Linguistics.

\bibitem[{Papineni et~al.(2002)Papineni, Roukos, Ward, and
  Zhu}]{papineni-etal-2002-bleu}
Kishore Papineni, Salim Roukos, Todd Ward, and Wei-Jing Zhu. 2002.
\newblock \href {https://doi.org/10.3115/1073083.1073135} {{B}leu: a method for
  automatic evaluation of machine translation}.
\newblock In \emph{Proceedings of the 40th Annual Meeting of the Association
  for Computational Linguistics}, pages 311--318, Philadelphia, Pennsylvania,
  USA. Association for Computational Linguistics.

\bibitem[{Ranasinghe et~al.(2020)Ranasinghe, Orasan, and
  Mitkov}]{ranasinghe-etal-2020-transquest}
Tharindu Ranasinghe, Constantin Orasan, and Ruslan Mitkov. 2020.
\newblock \href {https://doi.org/10.18653/v1/2020.coling-main.445}
  {{T}rans{Q}uest: Translation quality estimation with cross-lingual
  transformers}.
\newblock In \emph{Proceedings of the 28th International Conference on
  Computational Linguistics}, pages 5070--5081, Barcelona, Spain (Online).
  International Committee on Computational Linguistics.

\bibitem[{Rei et~al.(2020)Rei, Stewart, Farinha, and
  Lavie}]{rei-etal-2020-comet}
Ricardo Rei, Craig Stewart, Ana~C Farinha, and Alon Lavie. 2020.
\newblock \href {https://doi.org/10.18653/v1/2020.emnlp-main.213} {{COMET}: A
  neural framework for {MT} evaluation}.
\newblock In \emph{Proceedings of the 2020 Conference on Empirical Methods in
  Natural Language Processing (EMNLP)}, pages 2685--2702, Online. Association
  for Computational Linguistics.

\bibitem[{Sellam et~al.(2020)Sellam, Das, and Parikh}]{sellam-etal-2020-bleurt}
Thibault Sellam, Dipanjan Das, and Ankur Parikh. 2020.
\newblock \href {https://doi.org/10.18653/v1/2020.acl-main.704} {{BLEURT}:
  Learning robust metrics for text generation}.
\newblock In \emph{Proceedings of the 58th Annual Meeting of the Association
  for Computational Linguistics}, pages 7881--7892, Online. Association for
  Computational Linguistics.

\bibitem[{Takahashi et~al.(2020)Takahashi, Sudoh, and
  Nakamura}]{takahashi-etal-2020-automatic}
Kosuke Takahashi, Katsuhito Sudoh, and Satoshi Nakamura. 2020.
\newblock \href {https://doi.org/10.18653/v1/2020.acl-main.327} {Automatic
  machine translation evaluation using source language inputs and cross-lingual
  language model}.
\newblock In \emph{Proceedings of the 58th Annual Meeting of the Association
  for Computational Linguistics}, pages 3553--3558, Online. Association for
  Computational Linguistics.

\bibitem[{Vaswani et~al.(2017)Vaswani, Shazeer, Parmar, Uszkoreit, Jones,
  Gomez, Kaiser, and Polosukhin}]{vaswani2017attention}
Ashish Vaswani, Noam Shazeer, Niki Parmar, Jakob Uszkoreit, Llion Jones,
  Aidan~N Gomez, {\L}ukasz Kaiser, and Illia Polosukhin. 2017.
\newblock Attention is all you need.
\newblock \emph{Advances in neural information processing systems}, 30.

\bibitem[{Wan et~al.(2021)Wan, Liu, Yang, Bi, Zhang, Chen, Luo, Wong, and
  Chao}]{wan-etal-2021-robleurt}
Yu~Wan, Dayiheng Liu, Baosong Yang, Tianchi Bi, Haibo Zhang, Boxing Chen,
  Weihua Luo, Derek~F. Wong, and Lidia~S. Chao. 2021.
\newblock \href {https://aclanthology.org/2021.wmt-1.114} {{R}o{BLEURT}
  submission for {WMT}2021 metrics task}.
\newblock In \emph{Proceedings of the Sixth Conference on Machine Translation},
  pages 1053--1058, Online. Association for Computational Linguistics.

\bibitem[{Wan et~al.(2022{\natexlab{a}})Wan, Liu, Yang, Zhang, Chen, Wong, and
  Chao}]{wan-etal-2022-unite}
Yu~Wan, Dayiheng Liu, Baosong Yang, Haibo Zhang, Boxing Chen, Derek Wong, and
  Lidia Chao. 2022{\natexlab{a}}.
\newblock \href {https://doi.org/10.18653/v1/2022.acl-long.558} {{U}ni{TE}:
  Unified translation evaluation}.
\newblock In \emph{Proceedings of the 60th Annual Meeting of the Association
  for Computational Linguistics (Volume 1: Long Papers)}, pages 8117--8127,
  Dublin, Ireland. Association for Computational Linguistics.

\bibitem[{Wan et~al.(2020)Wan, Yang, Wong, Zhou, Chao, Zhang, and
  Chen}]{wan-etal-2020-self}
Yu~Wan, Baosong Yang, Derek~F. Wong, Yikai Zhou, Lidia~S. Chao, Haibo Zhang,
  and Boxing Chen. 2020.
\newblock \href {https://doi.org/10.18653/v1/2020.emnlp-main.80} {Self-paced
  learning for neural machine translation}.
\newblock In \emph{Proceedings of the 2020 Conference on Empirical Methods in
  Natural Language Processing (EMNLP)}, pages 1074--1080, Online. Association
  for Computational Linguistics.

\bibitem[{Wan et~al.(2022{\natexlab{b}})Wan, Yang, Wong, Chao, Yao, Zhang, and
  Chen}]{wan-etal-2022-challenges}
Yu~Wan, Baosong Yang, Derek~Fai Wong, Lidia~Sam Chao, Liang Yao, Haibo Zhang,
  and Boxing Chen. 2022{\natexlab{b}}.
\newblock \href {https://doi.org/10.1162/coli_a_00435} {Challenges of neural
  machine translation for short texts}.
\newblock \emph{Computational Linguistics}, 48(2):321--342.

\bibitem[{Wei et~al.(2022)Wei, Yu, Hu, Weng, Luo, and
  Jin}]{wei-etal-2022-learning}
Xiangpeng Wei, Heng Yu, Yue Hu, Rongxiang Weng, Weihua Luo, and Rong Jin. 2022.
\newblock \href {https://doi.org/10.18653/v1/2022.acl-long.546} {Learning to
  generalize to more: Continuous semantic augmentation for neural machine
  translation}.
\newblock In \emph{Proceedings of the 60th Annual Meeting of the Association
  for Computational Linguistics (Volume 1: Long Papers)}, pages 7930--7944,
  Dublin, Ireland. Association for Computational Linguistics.

\bibitem[{Wolf et~al.(2020)Wolf, Debut, Sanh, Chaumond, Delangue, Moi, Cistac,
  Rault, Louf, Funtowicz, Davison, Shleifer, von Platen, Ma, Jernite, Plu, Xu,
  Le~Scao, Gugger, Drame, Lhoest, and Rush}]{wolf-etal-2020-transformers}
Thomas Wolf, Lysandre Debut, Victor Sanh, Julien Chaumond, Clement Delangue,
  Anthony Moi, Pierric Cistac, Tim Rault, Remi Louf, Morgan Funtowicz, Joe
  Davison, Sam Shleifer, Patrick von Platen, Clara Ma, Yacine Jernite, Julien
  Plu, Canwen Xu, Teven Le~Scao, Sylvain Gugger, Mariama Drame, Quentin Lhoest,
  and Alexander Rush. 2020.
\newblock \href {https://doi.org/10.18653/v1/2020.emnlp-demos.6} {Transformers:
  State-of-the-art natural language processing}.
\newblock In \emph{Proceedings of the 2020 Conference on Empirical Methods in
  Natural Language Processing: System Demonstrations}, pages 38--45, Online.
  Association for Computational Linguistics.

\bibitem[{Yang et~al.(2021)Yang, Lei, Liu, Qi, and Lv}]{yang-etal-2021-pos}
Kexin Yang, Wenqiang Lei, Dayiheng Liu, Weizhen Qi, and Jiancheng Lv. 2021.
\newblock \href {https://doi.org/10.18653/v1/2021.acl-long.467}
  {{POS}-{C}onstrained {P}arallel {D}ecoding for {N}on-autoregressive
  {G}eneration}.
\newblock In \emph{Proceedings of the 59th Annual Meeting of the Association
  for Computational Linguistics and the 11th International Joint Conference on
  Natural Language Processing (Volume 1: Long Papers)}, pages 5990--6000,
  Online. Association for Computational Linguistics.

\bibitem[{Yang et~al.(2022)Yang, Liu, Lei, Yang, Zhang, Zhao, Yao, and
  Chen}]{yang-etal-2022-gcpg}
Kexin Yang, Dayiheng Liu, Wenqiang Lei, Baosong Yang, Haibo Zhang, Xue Zhao,
  Wenqing Yao, and Boxing Chen. 2022.
\newblock \href {https://doi.org/10.18653/v1/2022.findings-acl.318} {{GCPG}: A
  general framework for controllable paraphrase generation}.
\newblock In \emph{Findings of the Association for Computational Linguistics:
  ACL 2022}, pages 4035--4047, Dublin, Ireland. Association for Computational
  Linguistics.

\bibitem[{Yao et~al.(2020)Yao, Yang, Zhang, Chen, and
  Luo}]{yao-etal-2020-domain}
Liang Yao, Baosong Yang, Haibo Zhang, Boxing Chen, and Weihua Luo. 2020.
\newblock \href {https://doi.org/10.18653/v1/2020.coling-main.399} {Domain
  transfer based data augmentation for neural query translation}.
\newblock In \emph{Proceedings of the 28th International Conference on
  Computational Linguistics}, pages 4521--4533, Barcelona, Spain (Online).
  International Committee on Computational Linguistics.

\bibitem[{Zhang et~al.(2020)Zhang, Kishore, Wu, Weinberger, and
  Artzi}]{zhang2019bertscore}
Tianyi Zhang, Varsha Kishore, Felix Wu, Kilian~Q. Weinberger, and Yoav Artzi.
  2020.
\newblock \href {https://openreview.net/forum?id=SkeHuCVFDr} {Bertscore:
  Evaluating text generation with {BERT}}.
\newblock In \emph{8th International Conference on Learning Representations,
  {ICLR} 2020, Addis Ababa, Ethiopia, April 26-30, 2020}. OpenReview.net.

\bibitem[{Zhou et~al.(2020)Zhou, Yang, Wong, Wan, and
  Chao}]{zhou-etal-2020-uncertainty}
Yikai Zhou, Baosong Yang, Derek~F. Wong, Yu~Wan, and Lidia~S. Chao. 2020.
\newblock \href {https://doi.org/10.18653/v1/2020.acl-main.620}
  {Uncertainty-aware curriculum learning for neural machine translation}.
\newblock In \emph{Proceedings of the 58th Annual Meeting of the Association
  for Computational Linguistics}, pages 6934--6944, Online. Association for
  Computational Linguistics.

\end{thebibliography}
